\documentclass[a4paper,conference]{IEEEtran}
\usepackage{graphicx}
\usepackage{amsmath,amssymb} 
\usepackage{color}
\usepackage{algorithm}

\usepackage{adjustbox}
\usepackage{multirow}
\usepackage{float}
\usepackage{subfig}
\usepackage[noend]{algpseudocode}
\usepackage{wrapfig}
\usepackage{setspace}

\newcommand{\B}[1]{\textbf{#1}}
\newcommand{\I}[1]{\textit{#1}}

\newcommand{\s}[0]{\small}

\ifCLASSINFOpdf
\else
\fi
\begin{document}
%
\title{Beyond the Deep Metric Learning: Enhance the Cross-Modal Matching with Adversarial Discriminative Domain Regularization}



\author{\IEEEauthorblockN{Li Ren\IEEEauthorrefmark{1},
Kai Li\IEEEauthorrefmark{2},
LiQiang Wang\IEEEauthorrefmark{1},
Kien Hua\IEEEauthorrefmark{1}}

\IEEEauthorblockA{\IEEEauthorrefmark{1}University of Central Florida, \IEEEauthorrefmark{2}Facebook\\
renli@knights.ucf.edu, \{lwang, kienhua\}@cs.ucf.edu, kailee88@fb.com}
}

\maketitle

\begin{abstract}
Matching information across image and text modalities is a fundamental challenge for many applications that involve both vision and natural language processing. The objective is to find efficient similarity metrics to compare the similarity between visual and textual information. Existing approaches mainly match the local visual objects and the sentence words in a shared space with attention mechanisms. The matching performance is still limited because the similarity computation is based on simple comparisons of the matching features, ignoring the characteristics of their distribution in the data. In this paper, we address this limitation with an efficient learning objective that considers the \textit{discriminative feature distributions} between the visual objects and sentence words. Specifically, we propose a novel \textit{Adversarial Discriminative Domain Regularization} (ADDR) learning framework, beyond the paradigm metric learning objective, to construct a set of discriminative data domains within each image-text pairs. Our approach can generally improve the learning efficiency and the performance of existing metrics learning frameworks by regulating the distribution of the hidden space between the matching pairs. The experimental results show that this new approach significantly improves the overall performance of several popular cross-modal matching techniques (SCAN \cite{lee2018stacked}, VSRN \cite{li2019visual}, BFAN \cite{liu2019focus}) on the MS-COCO and Flickr30K benchmarks.
\end{abstract}


%
\IEEEpeerreviewmaketitle



\maketitle

\section{Introduction}

Techniques for analyzing data across multiple modalities bridge the vision and language areas, and play an essential role in many cross-modal applications such as image-text retrieval \cite{gu2018look,huang2018learning} and image captioning \cite{ren2018improved,ren2019improving}. In this paper, we focus on mapping the visual objects and sentence words into a shared hidden space to learn the similarity metric that compares the data from the different modalities. The fundamental challenge of cross-modal matching is to learn the ranked representations, in which the query sample should be closer to the matched sample than any other sample according to a certain distance metric. Specifically, those embedding parameters and similarity metrics are learned with the \textit{triplet loss} that contains a certain margin to optimize the data rankings \cite{faghri2017vse++,nam2017dual,niu2017hierarchical,huang2018learning}. While the highly informative region features can be extracted from a pre-trained Faster R-CNN \cite{ren2015faster} with the bottom-up attention approach (BU-ATT)  \cite{anderson2018bottom}, better retrieval results can be achieved with the learned representations that are embedded from advanced embedding networks and attention mechanisms\cite{lee2018stacked,li2019visual,liu2019focus,wehrmannadaptive}.


\begin{figure}[t]
    \centering
    \includegraphics[width=0.38\textwidth,height=0.18\textheight]{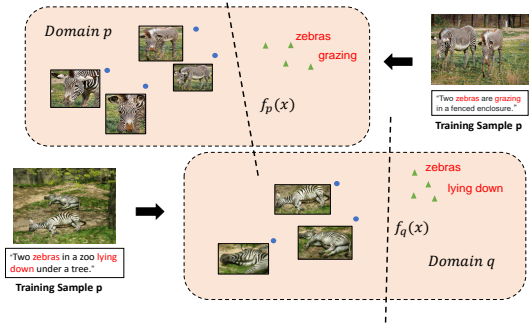}
    \caption{Illustrate an example how our regularization term help to distinguish the sample $p$ and $q$ that have very close semantic meaning.}
    \vspace{-10pt}
    \label{fig:introduction}
\end{figure}
\vspace{-2pt}

In order to efficiently search the features in hidden space across different modalities, some recent works also extend the metric learning objectives by adding a \textit{domain adaptation} learning task to align the feature distributions. The goal of the domain adaptation is to reduce the \textit{domain shift}, which is a classifier-induced divergence estimating the adaptability of the classifiers trained in the \I{source domain} and tested in the \I{target domain}. In order to feasibly estimate the divergence, one efficient way is to combine the domain adaptation with the adversarial learning techniques by jointly training the feature generators and the domain classifier (DANN) \cite{ganin2016domain,tzeng2017adversarial}. The cross-domain generators can then be learned in an adversarial way to confuse the domain classifier so that the distributions from the image and text domains are aligned.

The efficiency of applying domain adaptation in cross-modal matching tasks has been successfully reported in some recent research works \cite{wang2017adversarial,xu2018modal,peng2019cm,sarafianos2019adversarial}. Existing techniques, however, still have some significant limitations as follows:
\vspace{-2pt}
\begin{itemize}
	\item They focus on aligning all data samples into a single domain, in which the domain adaptation process rarely learn the diversity of semantic meaning among those samples. Some close visual objects or sentence terms that contain different global meanings are not obvious to be distinguished in this situation.
	\item The category knowledge in the image or the text domain is essential for existing methods in order to learn the \I{intro-modality} discriminative representation \cite{wang2017adversarial}. Existing approaches are not capable of learning enough discriminative information on the dataset without any category labels, e.g., MS-COCO, Flickr30K.
	\item Existing methods, based on matching the global visual representation, cannot take advantage of the most recent metric learning techniques that learn to aggregate the local visual objects, with the correlation and semantic reasoning among those objects, also considered. \cite{lee2018stacked,li2019visual,wehrmannadaptive}.
\end{itemize}
\vspace{-2pt}
To address these limitations, we propose to compare the local visual objects and textual words by generating discriminate representations with domain adaptation techniques to enhance the deep metric learning without referring to any category information. Instead of aligning all data samples into a unique domain, we separate distributions of each matching pair by training a group of independent discriminators. Specifically, we build a novel regularization framework to align the distribution of each correct matching image-text pair in the joint space with a collection of \I{discriminative feature domains}. In other words, we try to align the domains with similar semantic meanings while learning the distribution discrepancy between the samples that have different meanings. Moreover, we introduce an efficient \I{regularization term} to constructively restrict the discriminators in order to ensure that those independent classifiers describe the distinct space of classes.

Our overall intuition is based on the fact that the data samples from both image and text modality initially have the domain shift between each sample according to their semantic gap. We propose to learn the data distributions closely aligned with their semantic meanings by learning the domain invariant features with domain adaptation while keeping their discriminative semantic properties. We then try to regulate each domain's distinctiveness by comparing their domain classifiers with the classifiers of the other pairs. Our regularization term is compatible with the existing feature embedding strategies and the paradigm triple ranking losses. We demonstrate the efficiency of our regularization term base on three recent cross-modal matching frameworks, the SCAN \cite{lee2018stacked}, VSRN \cite{li2019visual} and BFAN \cite{liu2019focus}. Our experiments also show high learning efficiency and improved performance by evaluating them on two popular cross-modal matching datasets (MS-COCO and Flickr30k).

The effect of our discriminative regularization can be illustrated with an example in Fig. \ref{fig:introduction}. Assume we are comparing two image-text pairs which contain close semantic meaning that is easy to be mismatched. Our approach first adapts these pairs into two domains with two individual domain classifiers. To further increment their difference at the domain level, we compare the learning errors of these two classifiers in both image-text matching pairs. In each of the matching pairs, our regularization term tries to ensure that the actual classifier's prediction risk is less than the corresponding errors of the other classifiers. In a regularized domain space, the standing zebra is closer to the textual concepts (``grazing'', ``zebras'') than the (``lying down'', ``zebras'') concepts. They are difficult to be distinguished using vanilla metric learning approaches. Our example shows that the aligned discriminative domains help the metric learning model to search a hidden space where the features from different modalities with the same semantic meaning can be easily matched.

To sum up, the major contributions of this paper can be summarized as follows: 
\vspace{-2pt}
\begin{itemize}
	\item We propose a novel framework Adversarial Discriminative Domain Regularization (ADDR) that generally enhances the cross-modal metric learning networks. It is achieved by learning a group of discriminative domains regularized with a constructive learning term that explicitly aligned to each image-text pair.
	\item Our ADDR is compatible with existing metric learning networks. It is used as an add-on regularizer to their primary tasks to help match between a group of visual objects and the corresponding sentence.
	\item Our quantitative experiments show the effectiveness of our approach base on the recent popular metric learning frameworks: the SCAN \cite{lee2018stacked}, VSRN \cite{li2019visual}, and BFAN \cite{liu2019focus} on the popular MS-COCO and Flickr30k datasets.
\end{itemize}
\vspace{-2pt}
The remainder of this paper is organized as follows. We first discuss related works in Section \ref{related work}. Then in Section \ref{preliminaries}, we present a basic cross-modal matching solution and several existing architectures that we subsequently use as the backbone network to demonstrate and evaluate our approaches. The proposed independent domain adaptation and discriminative domain regulations are introduced in Section \ref{method}. Their effectiveness is evaluated in Section \ref{experiments}. Finally, we conclude this paper in Section \ref{conclusion}.

\section{Related Works}
\label{related work}

\subsection{Cross-Modal Matching}

The problem of matching and retrieving embedded samples cross the image and text modalities has been widely studied. Early works attempt to learn the alignment among image and text representations encoded with deep neural networks with a hinge-based triplet objective \cite{kiros2014unifying,karpathy2014deep}. Following those works, Nam et al. \cite{nam2017dual} match the textual sentence to an image that applies the attention mechanism on its sub-regions. Niu et al. \cite{niu2017hierarchical} propose a hierarchical RNN to learn the matching, that preserves the internal structure of sentences. Huang et al. \cite{huang2018learning} learn the sentence representation with a reconstruction loss to preserve the concept order of the sentence. Faghri et al. \cite{faghri2017vse++} further apply the hard negatives in the triplet loss to loosen its restriction and achieve significant improvement. Gu et al. \cite{gu2018look} propose the reconstruction of both image and text data with an adversarial training framework to enhance the semantic connection of the embeddings. Nikolaos et al. \cite{sarafianos2019adversarial} also propose to leverage the cross-entropy loss and adversarial training to estimate the matching scores.

\subsection{Visual Objects and Textual Words Matching}

Recently, Lee et al. \cite{lee2018stacked} model the visual and textual relationships with the help of local visual object features that are extracted from the Faster R-CNN pre-trained on a visual object dataset (Visual Genomes \cite{krishnavisualgenome}) with the bottom-up approach \cite{anderson2018bottom}. Several recent works also followed this pre-processing procedure to match the textual sentence with a group of local visual objects. Wang et al. \cite{wang2019position} combine the relative position and the visual features of those sub-region patches to boost the matching result. Hu et al. \cite{hu2019multi} model the relationships between the local regions and the words, and encode them with both visual and textual feature by two parallel attention networks. Li et al. \cite{li2019visual} leverage the Graph Convolutional Networks (GCN) combined with an RNN based walking model to learn the relationships among the visual objects while Chen et al. \cite{chen2020expressing} also encode the sub-region visual features with an RNN to model the intend correlations between the objects. Most recently, Jonatas et al. \cite{wehrmannadaptive} refine the adaptive visual and textual features to enhance their overall representative before comparing them in metrics.


\begin{figure*}[pt!]
	\centering
	\includegraphics[width=0.75\textwidth, height=0.26\textheight]{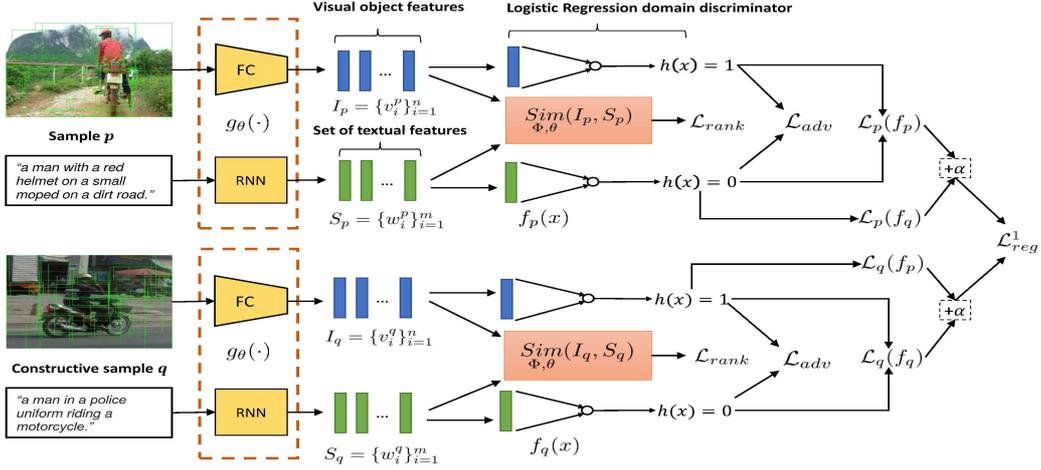}
	\caption{Illustrate the way that our ADDR objectives on sample p and q combined with existing metrics learning frameworks.}
	\vspace{-5pt}
	\label{framework_with_notation}
\end{figure*}


\section{Preliminaries}
\label{preliminaries}

In this section, we discuss the principle idea of the cross-modal matching task and introduce three existing successful deep metric learning frameworks that we will use for the evaluation of our approaches.

\subsection{Notation and Problem Definition}
The primary goal of cross-modal matching is to learn a deep similarity score metric with corresponding feature embedding networks to match the data from different modalities, that describe the same content. Specifically, the scope of this paper mainly focuses on the matching task involving visual objects in images and words in sentences. With the help of a pre-trained Faster-RCNN with the bottom-up attention (BU-ATT)\cite{ren2015faster,anderson2018bottom}, an image $I$ can be represented with a set of $n$  feature vectors $g_\theta(I) = \{v_i\}^n_{i=1}, v_i \in \mathbb{R}^D$ via a fully connected embedding layer $g(\cdot)_\theta$ with parameter $\theta$. Similarly, a sentence $S$ can also be represented as a set of $m$ feature vectors $g_\theta(S)=\{w_i\}^m_{i=1}, w_i \in \mathbb{R}^D$ that embedded with $g_\theta(\cdot)$ which is usually a bi-direction GRU embedding layer. Note that we denote the embedding networks (or the feature generators) for both domains as $g_\theta(\cdot)$ for convenience. In this paper, our goal is to learn a similarity-score function $Sim_{\Phi}(I, S)$ for matching $(I, S)$ pairs with corresponding metric parameters $\Phi$ and the optimized embedding networks $g_\theta(\cdot)$. We want to ensure that the measured similarity score between the positive (matched) image-text pair $(I, S)$ is higher than any other exclusive (mismatched) pairs, e.g., $(I, S')$ or $(I,' S)$ where $I \neq I'$ and $S \neq S'$. In the remainder of this paper, we also denote the similarity metric $Sim_\Phi$ with corresponding embedding networks $g_\theta(I), g_\theta(S)$ on pair $(I, S)$ as $\underset{\Phi, \theta}{Sim}(I, S)$ for convenience.

The current solution learns the mapping functions through the \textit{triplet ranking loss} $\mathcal{L}_{rank}$ that compares the positive pair with the negative pairs simultaneously. The intent of this loss is to ensure that the similarity scores of positive pairs are larger than the score of any other exclusive pairs with a particular value of margin. Faghri et al. \cite{faghri2017vse++} also suggest to separate the most confusing mismatching pairs (also called \textit{hard negative pairs}) instead of all negative pairs. One specific triplet ranking loss in cross-modal metric learning can be expressed as:

\vspace{-2pt}
\begin{equation}
\begin{split}
\mathcal{L}_{rank}(I, S) & = \max[0, \delta - \underset{\Phi, \theta}{Sim}(I, S) + \underset{\Phi, \theta}{Sim}(I, \hat{S})] \\
& + \max[0, \delta - \underset{\Phi, \theta}{Sim}(I, S) +  \underset{\Phi, \theta}{Sim}(\hat{I}, S)],
\label{instance_loss}
\end{split}
\end{equation}
\vspace{-2pt}
where $\hat{S}=argmax_{S' \neq S}\underset{\Phi, \theta}{Sim}(I, S')$ and $\hat{I}=argmax_{I' \neq I}\underset{\Phi, \theta}{Sim}(I', S)$, which indicate the hard negative pairs within all exclusive negative pairs $(I', S)$ and $(I, S')$. Here we use the three most recent successful architectures, namely SCAN, VSRN, and BFAN, to estimate $Sim_{\Phi}(I, S)$ with the corresponding parameters $\Phi$. They match visual objects and words in sentences providing the backbone metric learning frameworks to evaluate our method.

\section{Proposed Method}
\label{method}

In this section, we present the general methodology and discuss the intuition of our approach in a domain adaptation perspective. As discussed above, our goal is to leverage the metrics learning frameworks to learn better-ranked similarity metrics $Sim_\Phi(I, S)$. To achieve this, we propose to enhance the representation generators $g_\theta(\cdot)$ to generate \I{domain invariant} representations to optimize the learning process of the similarity metrics $Sim_{\Phi}$. Specifically, we study the feature structure and learn to adapt the \textit{feature domains} with an additional adversarial loss $\mathcal{L}_{adv}(\cdot)$ to adapt the feature domains and a regularization loss $\mathcal{L}_{reg}(\cdot)$ constrain the discriminative domain structures. Intuitively, we leverage the connection between the semantic meaning of each image-text pairs and their feature distributions to ensure the data distributions within the positive pair $(I, S)$ are aligned together while separating it from the distributions of other pairs.

\subsection{Cross-Modal Matching with Domain Adaptation}
\label{DDA}


In this paper, we consider to train the feature generator $g_\theta(\cdot)$ by solving an \textit{unsupervised domain adaptation} problem, which aims to generate the features with less domain divergence from both image and text data in order to support the metric learning process. Generally, for generated image features $g_\theta(I)=\{v_i\}_{i=1}^n$ and text features $g_\theta(S)=\{w_i\}_{i=1}^m$, the goal of domain adaptation is to learn the \I{domain invariant} features where we could easily find a classifier $f(\cdot)$ that has low prediction risk on the label set $\mathcal{Y} = \{y_i\}_{i=1}^n$ in both image and text domains. Specifically, we follow the principle idea of adversarial domain adaptation (DANN) \cite{ajakan2014domain,ganin2016domain} where they estimate the domain divergence by learning a domain predictor. Thus in positive pair $p$, we can learn a classifier $f_p(\cdot)$ that minimizes the corresponding domain prediction risk. 




Note that we assume the domain classifier $f_p$ and the feature generator $g_\theta(\cdot)$ are trained with individual class labeling space of each image-text pair $(I_p, S_p)$ independently. We denote the adapted feature $x$ that is originally from the image domain in pair $p$ as $v^p$ and the corresponding feature from the text domain as $w^p$. Since our labeling space is symmetric, we would also assume the features $v^p$ is labeled as $h(v^p)=1$ and $w^p$ is labeled as $h(w)=0$ for all $N$ matching pairs in our training set for convenience. Moreover, we estimate the classifier $f_p(\cdot)$ as a single \textit{Logistic Regression} layer as suggested in \cite{ajakan2014domain, ganin2016domain}. In order to search the $f_p$ with a low risk of $\mathcal{R}_{D_p}(f_p)$, we optimize the cross-entropy objective $\mathcal{L}_{adv}$ for pair $p$ as following:
\vspace{-2pt}
\begin{equation}
\begin{split}
\min_{W_p, b_p} \mathcal{L}_{adv}(I_p, S_p) & = \sum^n_{i=1} log(\sigma(W_p^T w^p_i + b_p)) \\
			& + \sum^m_{i=1} log(1 - \sigma(W_p^T v^p_i  + b_p)),
\label{domain_g}
\end{split}
\end{equation}
where $\sigma$ represents the Sigmoid function and the parameter set $W_p \in \mathbb{R}^D, b_p \in \mathbb{R}$ are the parameters of the domain discriminator $f_p$. Here we use $\{W, b\}  = \{W_p, b_p\}_{p=1}^N$ to denote the set of parameters of those discriminators.

We then constrain the feature generators $g_\theta(\cdot)$ to minimize the divergence between $g_\theta(I_p)$ and $g_\theta(S_p)$ where we empirically maximize the lower bound of $\mathcal{R}_{D_p}(f_p)$. Thus, the objective to update the parameters of embedding network $\theta$ and the downstream similarity metric $\Phi$ can be expressed as follows:
\begin{equation}
\min_{\Phi, \theta}\frac{1}{N}\sum^N_{p=1}[\mathcal{L}_{rank}(I_p, S_p) - \beta\mathcal{L}_{adv}(I_p, S_p)],
\label{domain_d}
\end{equation}
where $L_{rank}(I_p, S_p)$ is the triplet loss for each positive pair $p$ described in Eq. \ref{instance_loss}; $\mathcal{L}_{adv}(I_p, S_p)$ is the corresponding entropy loss introduced in Eq. \ref{domain_g}, and $\beta$ is a hyperparameter that scale the adversarial objective. In the training procedure, we alternatively optimize the set of parameters $\mathcal{W}$ for domain discriminators and $\Phi, \theta$ for the feature generators and the downstream similarity metric, so that the feature generators $g_\theta(\cdot)$ can be learned in a trade-off between ranking similarities of the samples and minimizing the divergences between the image-text domain.

\subsection{Discriminative Domain Regularization}

\label{DDR}
Since $\mathcal{L}_{adv}$ introduced above identically construct labeling functions in hypothesis class space instead of real category labels, it is essential to constrain the searching process of $f_p$ in order for them to estimate the \textit{discriminative} class spaces. In other words, assuming we have a matching pair $(I_p, S_p)$ and an exclusive matching pair $(I_q, S_q)$, the class labeling space in the domain $p$ and $q$ may not be fully disjoint during the training. The triplet loss indeed provides some discriminative learning signals to feature generators, but only at the feature level. At the domain level, the $f_p$ and $f_q$ might still send confusing learning signals, and it may fail to capture the discriminative information between the distribution of samples.

To address this issue, we further introduce a regularization term to force each independent discriminators to become distinct to others in order to separate the feature spaces. Assuming we have a matching pair $(I_p, S_p)$ with its discriminator $f_p(\cdot)$ as discussed above, we can select sample pairs $(I_q, S_q)$ and $(I_r, S_r)$, where $S_q$ is a \I{hard negative} sample (most confusing sample selected by the similarity metric) to $I_p$ and $I_r$ is the corresponding hard negative sample to $S_p$. Also, we can correspondingly search discriminators $f_q(\cdot)$ and $f_r(\cdot)$ to separate the feature domains in data pair $(I_q, S_q)$ and $(I_r, S_r)$.




We intend to learn domain predictors in separated class space in domain $p$,  $q$ and $r$, and also learn discriminative domain invariant features under those spaces. Formally, for positive data pair $p$ and negative data pairs $q$ and $r$, we would constrain their risks as follows,
\vspace{-2pt}
\begin{align}
&\mathcal{R}_{D_p} (f_p) \leq \mathcal{R}_{D_p} (f_q) + \alpha \\
&\mathcal{R}_{D_p} (f_p) \leq \mathcal{R}_{D_p} (f_r) + \alpha \\
&\mathcal{R}_{D_q} (f_q) \leq \mathcal{R}_{D_q} (f_p) + \alpha \\
&\mathcal{R}_{D_r} (f_r) \leq \mathcal{R}_{D_r} (f_p) + \alpha 
\end{align}
where $\alpha$ is a pre-defined constrain margin and $\mathcal{R}_{D_p} (f_p)$ represents the prediction risk of $f_p$ in data pair $p$ with distribution $D_p$ as discussed in Section \ref{DDA}. Then, the regularization term $\mathcal{L}_{reg}^1$ that compares the pair $(I_p, S_p)$ and $(I_q, S_q)$ and the regularization term $\mathcal{L}_{reg}^2$ that compares the pair $(I_p, S_p)$ and $(I_r, S_r)$ can be expressed as follows:

\vspace{-4pt}
\begin{equation}
\begin{split}
\mathcal{L}_{reg}^1(I_p, S_p, I_q, S_q) &= \max[0, \alpha + \mathcal{L}_p (f_p) - \mathcal{L}_p (f_q),] \\ 
							  &+ \max [0, \alpha + \mathcal{L}_q (f_q) - \mathcal{L}_q (f_p)]						  
\label{domain_const_1}
\end{split}
\end{equation}
\vspace{-2pt}
\begin{equation}
\begin{split}
\mathcal{L}_{reg}^2(I_p, S_p, I_r, S_r) &=\max [0, \alpha + \mathcal{L}_q (f_q) - \mathcal{L}_q (f_p)] \\
							 					     &+ \max[0, \alpha + \mathcal{L}_p (f_p) - \mathcal{L}_p (f_r)],
\label{domain_const_2}
\end{split}
\end{equation}
where $\mathcal{L}_p (f_p)$ is the regression loss with parameter $W_p, b_p$ with labels $1, 0$ as we introduced in Section \ref{DDA}, the same with other losses. The overall regularization term $\mathcal{L}_{reg}$ can be defined as the sum of $\mathcal{L}_{reg}^1$ and $\mathcal{L}_{reg}^2$ as follows,
\vspace{-2pt}
\begin{equation*}
\mathcal{L}_{reg}(p, q, r) = \mathcal{L}_{reg}^1(I_p, S_p, I_q, S_q) + \mathcal{L}_{reg}^2(I_p, S_p, I_r, S_r)
\end{equation*}

In short, for any positive sample pair $(I_p, S_p)$, we compare the discriminator $f_p$ with the $f_q$ and $f_r$, which are the discriminators of its negative pairs. We optimize the generators accordingly to ensure that the class space in the domain $p$ is distinct from $q$ and $r$ in the \I{discriminator training phase} while the features in $(I_p, S_p)$ are distinct from $(I_q, S_q)$ and $(I_r, S_r)$ in the \I{generator training phase}.

\vspace{2pt}
\subsection{The Learning Objective and Optimization}

Integrating the objectives introduced above gives our overall learning objective $\mathcal{L}_{ADDR}^d$ in discriminator training phase and $\mathcal{L}_{ADDR}^g$ in generator training phase as follows:
\vspace{-2pt}
\begin{equation}
\mathcal{L}_{ADDR}^d = \frac{1}{N}\sum^N_{p=1} \{ \mathcal{L}_{adv}(I_p, S_p) + \gamma \mathcal{L}_{reg}(p, q, r)\} 
\label{overall_loss_d}
\end{equation}
\vspace{-2pt}
\begin{equation}
\mathcal{L}_{ADDR}^g = \frac{1}{N}\sum^N_{p=1} \{\mathcal{L}_{rank}(I_p, S_p) - \beta \mathcal{L}_{adv}(I_p, S_p) \}
\label{overall_loss_g}
\end{equation}
\vspace{-2pt}
where $\mathcal{L}_{rank}$ is the original triplet ranking loss of Eq. \ref{instance_loss} that optimizes the parameters of similarity metric $\Phi$ and the parameters of embedding networks $\theta$. $\mathcal{L}_{adv}$ is the adversarial object introduced in Eq. \ref{domain_d} and $\mathcal{L}_{reg}(p, q, r)$ is the regularization term introduced in Eq. \ref{domain_const_1}, Eq. \ref{domain_const_2}. $\beta$ and $\gamma$ are the pre-defined hyperparameters that empirically scale and balance the learning signals from $\mathcal{L}_{rank}$, $\mathcal{L}_{adv}$, and $\mathcal{L}_{reg}$. 

Different from existing adversarial cross-modal matching frameworks \cite{wang2017adversarial,xu2019deep,peng2019cm,sarafianos2019adversarial}, our framework intend to optimize a subset of the domain classifier parameters $\{W, b\} = \{W_p, b_p\}_{p=1}^N$ for each iteration while updating the whole encoder parameters $\theta$ and metric parameters $\Phi$. Thus in each epoch, the distribution of the embedded features will become unbalanced between each mini-batch. In order to solve this and balance the training signals for all N domain predictors, we separate the progress of updating parameters $\{W, b\}$ from updating $\theta$ and $\Phi$. In other words, we train the domain classifier parameters in different epoch from the one training the encoder and the similarity metric. The details of the training process of our ADDR is presented in Algorithm \ref{training}.

\begin{algorithm}
	\small
	\caption{Adversarial Discriminative Domain Regularization (ADDR)}
	\label{training}
	\begin{spacing}{1.0}
		\begin{algorithmic}[1]
			\State \textbf{Input: } Training Set $\mathcal{Q}={(I_p, S_p)}_{p=1}^N$ with raw image features and sentence terms.
			Hyperparameters $\delta, \alpha, \beta, \gamma$.
			\State \textbf{Output: } Learned Parameters $\theta$, $\Phi$
			\State \textbf{Initial: } $\theta, \Phi, \mathcal{W} = \{w_p, b_p\}_{p=1}^N$
			\While {stop criteria is not satisfied}
			\State /* \I{Discriminator Training Phase} begin */
			\For {each \textit{mini-batch} of size $k$} 
				\State Select data $I = {(I_p)}_{p=1}^{k}, S = {(S_p)}_{p=1}^{k}$
				\State Select Parameters $W = \{W_p\}_{p=1}^k$ and $b = \{b_p\}_{p=1}^{k}$
				\State Embedding $\{v^p\}_{p=1}^{k}\gets g_\theta(I)$, $\{w^p\}_{p=1}^{k} \gets g_\theta(S)$
				\State Calculate Metric Scores $ \mathcal{S} = \{Sim_{\Phi}(I_p, S_p)\}_{p=1}^{p={k}}$
				\State Select \I{hard negative} samples $(I_q, S_q)$ and $(I_r, S_r)$  
				\State Calculate $\Delta W, \Delta b  \gets \frac{\partial{\mathcal{L}_{adv}}}{W, b} + 
					\gamma ( \frac{\partial{\mathcal{L}_{reg}^1}}{W, b} + \frac{\partial{\mathcal{L}_{reg}^2}}{W, b}) $
				\State Update $W,  b \gets W,  b - Adam(\Delta W, \Delta b)$
			\EndFor
			\State /* \I{Generator Training Phase} begin */
			\For {each \textit{mini-batch} of size $k$}
					\State Repeat L7 - L10
					\State Calculate $\Delta \Phi \gets \frac{\partial{\mathcal{L}_{rank}}}{\Phi}$
					\State Calculate $\Delta \theta \gets \frac{\partial{\mathcal{L}_{rank}}}{\theta} - \beta \frac{\partial{\mathcal{L}_{adv}}}{\theta}$
					\State Update $\Phi \gets \Phi - Adam(\Delta \Phi)$
					\State Update $\theta \gets \theta - Adam(\Delta \theta)$
			\EndFor
			\EndWhile
		\end{algorithmic} 
	\end{spacing}
\end{algorithm}


\vspace{-2pt}
\section{Experiments}
\label{experiments}
The MS-COCO and Flickr30K dataset have been widely used in existing cross-modal retrieval works \cite{lee2018stacked,faghri2017vse++,zheng2017dual,gu2018look,huang2018learning,wang2019matching, hu2019multi, li2019visual}. In this section, we also evaluate our proposed framework on these two popular datasets.
\vspace{-2pt}
\subsection{Data and Training Setup}

\subsubsection{Datasets}
We first evaluate the performance of our method on the MS-COCO dataset, which contains 123,287 image samples, each labeled with 5 text descriptions. We follow the existing works \cite{zheng2017dual,lee2018stacked,faghri2017vse++} to set 11,328 images for training, 1000 for validation, and 5000 for testing. We also test it on Flickr30K, a smaller dataset with 31,000 images downloaded from the Flickr Website. Similar to MS-COCO, each image sample also has 5 corresponding sentence labels. We also follow the splitting in previous works to sample 1,000 images for validation, 1,000 for testing, and the rest of them are for training. For visual objects in both datasets, we follow the procedure provided by \cite{lee2018stacked} to extract 36 visual object features as with dimension size 2048 as our visual feature input.

\subsubsection{Network Setting}

To fairly compare with the recent approaches, we follow the original architectures and corresponding hyper-parameters proposed in SCAN, VSRN, and BFAN, including the feature and embedding dimensions in each network layer, the training batch size, and other structure-related parameters. We only replace the training strategy for our ADDR algorithm, as proposed in Algorithm \ref{training}. Specifically, the text encoder for SCAN and BFAN is a bi-directional GRU network; and for VSRN, we adopt the single layer GRU network they propose in the paper. The image encoder is a fully connected layer (FC), as suggested in their papers. 

Note that we train 113,287 discriminators for MSCOCO, and 29,000 discriminators for Flickr30K. In order to save memory resource, we put our discriminators in CPU memory instead of GPU. Our discriminators in MSCOCO take 1.8Gb, and in Flickr30K take 454Mb CPU memory in total. We train our model in a machine with E5-2640v4 CPU and 1080ti GPU. 
\begin{table}[ht]
	\small
	\centering
	\begin{adjustbox}{max width=0.8\textwidth}
	\begin{tabular}{|c|c|c|c|} \hline 
		
		  Methods  & R@1(i2t)    &  R@1(t2i)  &   Sum(ALL)  \\  \hline
		 \multicolumn{4}{|c|}{ADDR-SCAN}  \\  \hline
		SCAN \cite{lee2018stacked} 	& 70.9	&  56.4    &  500.5	\\ \hline
		Base-SCAN  						&  70.3	& 56.8 	& 500.6	\\ \hline
		United-SCAN & 72.3 	& 58.1  	& 504.1	 \\ \hline
		Multiple-SCAN &  72.6	 & 58.3  	&  505.7  \\ \hline 
		ADDR-SCAN  &  \B{74.8}   & \B{60.1} 	& \B{509.5}   \\ \hline 
		
		 \multicolumn{4}{|c|}{ADDR-BFAN} \\  \hline		  
		BFAN \cite{liu2019focus}  		&  73.7	&   58.3  &  -- \\ \hline
		Base-BFAN   						&  73.1	&  58.0	& 504.8	\\ \hline
		United-BFAN    							&  73.6	 & 58.6  	& 506.4	 \\ \hline 
		Multiple-BFAN     						&  73.7	 & 58.4  	&  506.9  \\ \hline 
		ADDR-BFAN &  \B{74.1}   &  \B{59.3} &  \B{511.1}  \\ \hline
		
		\multicolumn{4}{|c|}{ADDR-VSRN} \\  \hline
		 VSRN \cite{li2019visual} & -- 		&  -- 	&  --	\\ \hline
		Base-VSRN					    & 73.2  	&  59.6 & 506.6 \\ \hline
		United-VSRN    					&  74.0	 &  60.2 	& 509.1	 \\ \hline 
		Multiple-VSRN     				&  74.2	 &  59.8 	& 509.7   \\ \hline 
		ADDR-VSRN  					    &  \B{75.1}  &  \B{60.7} 	& \B{513.6} \\ \hline 
	\end{tabular}
	\end{adjustbox}
	\vspace{3pt}
	\caption{Ablation Studies with 3 backbone models on MS-COCO 1K test in 5-folder cross validation.}
	\vspace{-5pt}
	\label{ablation}
\end{table}

\begin{table*}[t]
	\small
	\centering
	\begin{adjustbox}{max width=0.9\textwidth}
	\begin{tabular}{|c|c|c|c|c|c|c|c|} \hline
		& \multicolumn{3}{|c|}{Sentence Retrieval} & \multicolumn{3}{|c|}{Image Retrieval}  & \multirow{2}{*}{Sum (ALL)} \\ 
		\cline{1-7}
		Method  & R@1    & R@5     & R@10     & R@1     & R@5     & R@10  & \\  \hline
		\multicolumn{8}{|c|}{1k Test Set (5-fold)} \\ \hline 
		
		SCAN \cite{lee2018stacked} \s{(2018)}  				& 72.7  		& 94.8  	& 98.4  	& 58.8  	& 88.4  & 94.8 					& 507.9 		\\ \hline
		MTFN \cite{wang2019matching} \s{(2019)}		& 74.3		& 94.9	& 97.9		& 60.1		& 89.1	& 95.0 		& 511.3		\\ \hline
		BFAN \cite{liu2019focus} \s{(2019)} 					& 74.9 		&	95.2 	& 98.3		& 59.4		& 88.4	& 94.5 		&  510.7		\\ \hline		
		VSRN \cite{li2019visual} \s{(2019)}						& 76.2 		& 94.8 		& 98.2 		& 62.8 		& 89.7 	& 95.1 	& 516.8  	\\ \hline 
		DPRNN \cite{chen2020expressing} \s{(2020)}	& 75.3		& 95.8		& 98.6		& 62.5		& 89.7	& 95.1 	& 517.0		\\	\hline
		ADAPT \cite{wehrmannadaptive} \s{(2020)}		& 76.5		& 95.6		& 98.9		& 62.2		& 90.5	& 96.0 	& 519.7		\\ \hline 
		\hline
		ADDR-SCAN \s{(Ours)}  &  76.1  			& 95.5  		&  98.4 	&  61.2 			&  88.9			&  94.8	& 514.9	\\ \hline
		ADDR-BFAN \s{(Ours)}  &   76.4  			&  95.8 		&  98.3 	&  62.3 			&  89.4 			& 96.2 	&  518.4	\\ \hline
		ADDR-VSRN \s{(Ours)}   &   \B{77.4} 	& \B{96.1}  & \B{98.9}  	&  \B{63.5}   	&  \B{90.7} 	& \B{96.7}  & \B{523.3}	\\ \hline 
		
		\multicolumn{8}{|c|}{5K Test Set} \\ \hline 
		
		SCAN \cite{lee2018stacked} \s{(2018)}			& 50.4  & 82.2   & 90.0    		& 38.6    	& 69.3    & 80.4 &  410.9	\\ \hline 
		MTFN \cite{wang2019matching} \s{(2019)}	& 48.3	& 77.6	 & 87.3		& 35.9		& 66.1 	& 76.1 &  391.3	\\ \hline 
		BFAN \cite{liu2019focus} \s{(2019)}				& 52.9  & 82.8		& 90.6		& 38.3		& 67.8	& 79.3 & 411.7	\\ \hline
		VSRN \cite{li2019visual} \s{(2019)} 				& 53.0	& 81.1	 & 89.4		& 40.5		& 70.6	& 81.1 &  415.7	\\ \hline
		\hline
		ADDR-SCAN \s{(Ours)}  &  \B{57.3}  	& \B{86.0} &  \B{92.7}&  41.8 			&  \B{72.0} 	& 81.3 		& 	\B{431.1}	\\ \hline
		ADDR-BFAN \s{(Ours)}   &   54.3 			& 84.0   		&	 91.5 		&  40.1 			&  69.2  			& 80.6  		&		419.7	\\ \hline 
		ADDR-VSRN \s{(Ours)}   &   56.6  		&  85.3 		&  90.4 		&   \B{42.5} 	&  71.9 			& \B{82.0} &		428.7	\\ \hline 
		
	\end{tabular}
	\end{adjustbox}
	\caption{Compare the Retrieval Result on the test set of MS-COCO from ensemble results (sorted by publishing date).}
	\vspace{-10pt}
	\label{mscoco}
\end{table*}

\begin{table*}[h]
	\small
	\centering
	\begin{adjustbox}{max width=0.9\textwidth}
	\begin{tabular}{|c|c|c|c|c|c|c|c|} \hline 
		& \multicolumn{3}{|c|}{Sentence Retrieval} & \multicolumn{3}{|c|}{Image Retrieval}  & \multirow{2}{*}{Sum (ALL)} \\ 
		\cline{1-7}
		Method  & R@1    & R@5     & R@10     & R@1     & R@5     & R@10  & \\  \hline
		SCAN \cite{lee2018stacked} \s{(2018)} 				& 67.4    	& 90.3  	& 95.8  		& 48.6  	& 77.7  & 85.2 	& 465.0	\\ \hline
		MTFN \cite{wang2019matching} \s{(2019)}		& 65.3	  	& 88.3	& 93.3		& 52.0	& 80.1	& 86.1 & 465.1	\\ \hline
		BFAN \cite{liu2019focus} \s{(2019)} 					& 68.1 		& 91.4  	& 95.9		& 50.8	& 78.4 	& 85.8 & 470.4	\\ \hline	
		VSRN \cite{li2019visual} \s{(2019)} 					& 71.3	  	& 90.6	& 96.0		& 54.7	& 81.8	& 88.2 & 482.6	\\ \hline
		RDAN \cite{hu2019multi}\s{(2019)}					& 68.1		& 91.0	& 95.9		& 54.1	& 80.9	& 87.2 & 477.2	\\ \hline
		\hline
		ADDR-SCAN \s{(Ours)}   &   72.1  	&   \B{93.1}		&  96.1   	&   53.5  &  80.4   &	 87.4	& 482.6 \\ \hline
		ADDR-BFAN \s{(Ours)}  	&   71.3		&   91.5    			&  96.4     	&   54.0	&  80.0 	&  87.6	&  480.8 \\ \hline 
		ADDR-VSRN \s{(Ours)}  	&    \B{73.0} &    92.5   	&  \B{96.6}    &   \B{55.6}	&  \B{82.0} 	& \B{88.9}	& \B{488.6} \\ \hline 
	\end{tabular}
	\end{adjustbox}
	\caption{Comparison of the Retrieval Results of the ensamble results on the test set of Flickr30K.}
	\vspace{-10pt}
	\label{flickr30k}
\end{table*}

\subsubsection{Ablation Study}
To evaluate the component of our regularization terms individually, we do the ablation study on MS-COCO 1k test set to compare their performance in all three backbone architectures. Our baseline (labeled as Base) is built according to the settings of \I{single model} in corresponding papers (e.g., single SCAN-t2i, single model VSRN, and single BFAN-prob). For each setting, we follow the same training strategy to evaluate all objective terms ($\mathcal{L}_{rank}, \mathcal{L}_{adv}, \mathcal{L}_{reg}$). We also evaluate the existing domain adaptation strategy for cross-modal matching, where all sample pairs are adopted into a united domain (labeled as United), which is most close to the existing adversarial cross-modal matching approaches \cite{wang2017adversarial,sarafianos2019adversarial,peng2019cm,sarafianos2019adversarial}. The detail of the ablation setting and performance can be regarded in table \ref{ablation}.

\subsubsection{Evaluation metric}
To evaluate the learning effect of the metric, we adopt the task of \I{sentence retrieval} and \I{image retrieval}, in which we measure the performance by calculating its recall at top $k$ values as $R@k$ ($k=1,5,10$), which is consistent with previous works. They are defined as the fraction of queries for which the correct item is retrieved in the closest $k$ samples. We also evaluate the overall performance by calculating the sum of all 6 recall scores, as suggested in other existing works.

\vspace{-2pt}
\begin{figure}[pt]
\subfloat[\label{subfig-1}]{
		\includegraphics[width=0.48\textwidth,height=0.14\textheight]{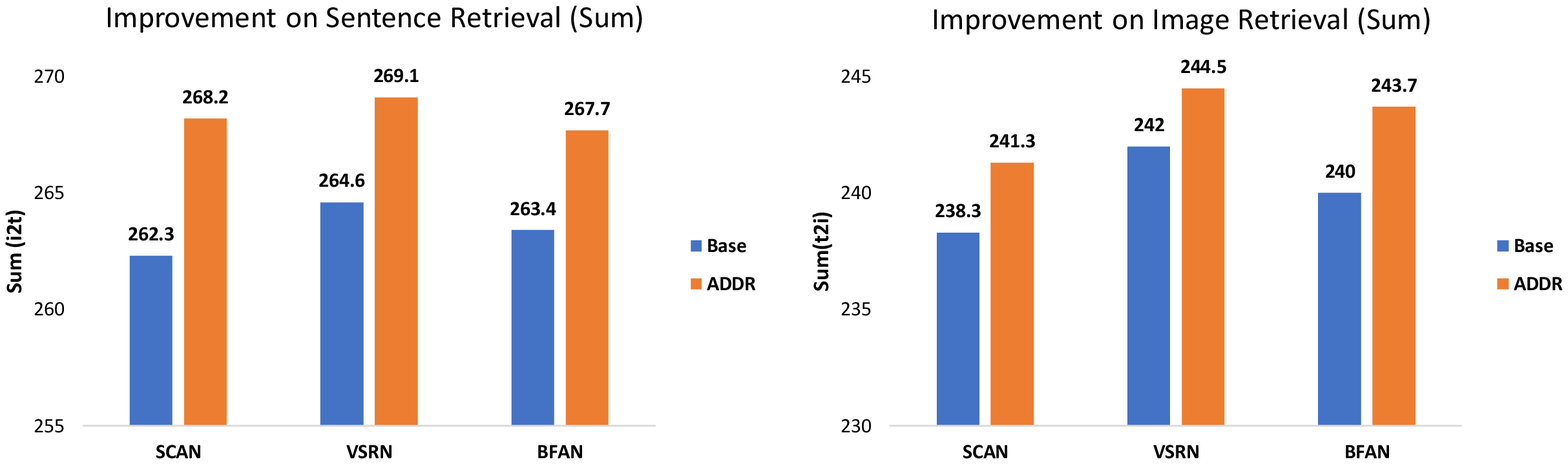}
	}
	\newline
\subfloat[\label{subfig-2}]{
		\includegraphics[width=0.45\textwidth,height=0.16\textheight]{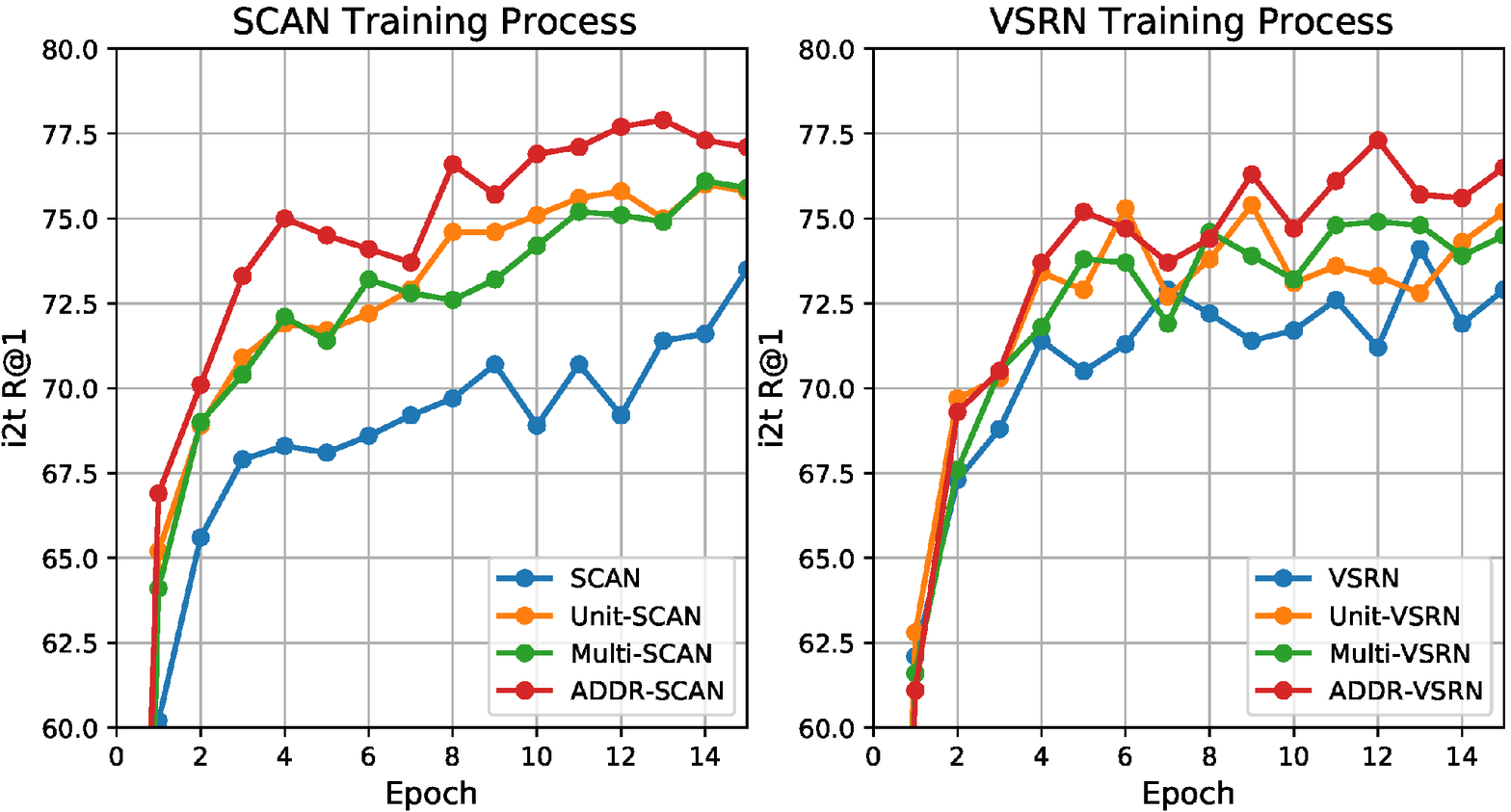}
	}
	\caption{(a) Compares the sum scores between our ADDR and baseline, and (b) plots the R@1 score on validation set during training phase.}
	\vspace{-5pt}
	\label{improvement}
	\vspace{-10pt}
\end{figure}

\begin{figure*}[ht]
\includegraphics[width=0.85\textwidth,height=0.20\textheight]{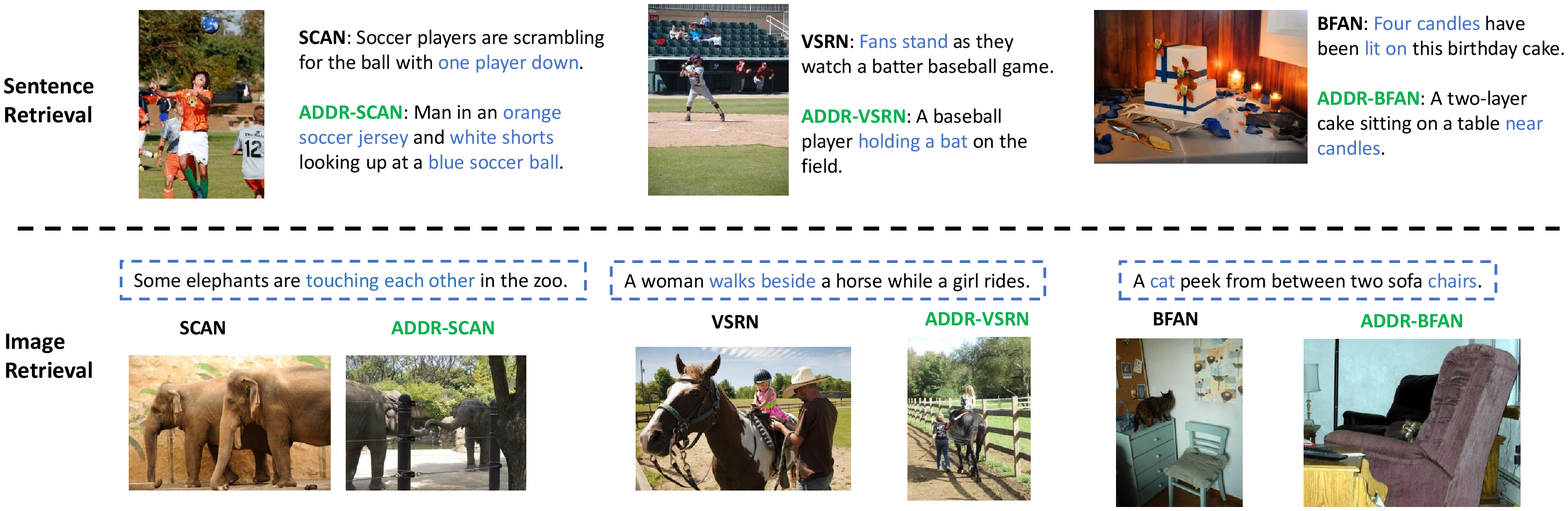}
\caption{Selected examples of qualitative image retrieval and sentence retrieval on MS-COCO dataset. We highlight the discriminate concepts or phrase that the model learn to distinguish from other confusing samples. Green name is the method retrieve the correct results in our experiments.}
\label{examples}
\vspace{-5pt}
\end{figure*}

\vspace{2pt}
\subsubsection{Training Profile and Parameter Searching}

For both of the feature generator and domain discriminator, we adopt the Adam optimizer ($\beta_1=0.5, \beta_2=0.999$) \cite{kingma2014adam} that is efficient to optimize the general embedding networks. For all experiments, we fix the batch size to 128, which is consistent with their original papers. We train the three models with the learning rate 0.001 for SCAN, 0.0005 for BFAN, and 0.0002 for VSRN correspondingly.

In the training phase, we manually decay the learning rate by ratio 0.2 for every 8000 iterations until it reaches $10^{-5}$. In each iteration, we select a subset of parameters $\{w_1,\dots w_k\}$ for discriminators from a hash table of the parameter set $\mathcal{W}$ according to the index of training samples. We then continue to train the selected discriminators with one image sample and 5 of its paired sentences. We search our hyper-parameters $\{\beta, \gamma\}$ from a grid where for $\beta, \gamma$ we search in a ranges from $10^{-3}$ to $1.0$, and the best performance is achieved when $\{ \beta=0.1, \gamma=0.4\}$, $\{\beta=0.1, \gamma=0.7 \}$, $\{\beta=0.1, \gamma=0.1 \}$ for SCAN, VSRN and BFAN respectively. For the margin $\alpha$ to compare the domain predictors, we empirically select 0.05 in all our experiments.

\vspace{-5pt}
\subsection{Qualitative Results}

We list the single model results of the ablation study on 1k datasets of MS-COCO in table \ref{ablation}. In comparison with the performance of our base model (Base-SCAN, Base-BFAN, and Base-VSRN) and the reference model with single domain adaptation (Unite-SCAN, Unite-BFAN, and Unite-VSRN), this study indicates that our baseline is reliable and comparable to the reported value from the original papers. We conclude that the domain adaptation technique is generally effective on all backbone models. We notice that applying multiple domain adaptations without any regularization process (Multi-SCAN, Multi-BFAN, and Multi-VSRN) leads to results close to the single domain adaptation settings, as discussed in Section \ref{DDR}. Specifically, our proposed running (ADDR-SCAN, ADDR-BFAN, and ADDR-VSRN) outperforms our baseline and other ablation settings in both image and sentence retrieval tasks.

Figure \ref{improvement} (a) also illustrates the comparison of text retrieval (i2t) and image retrieval (t2i) results separately. We also study the learning efficiency by comparing the R@1 scores in training, as illustrated in Figure \ref{improvement} (b). We find that ADDR guides the optimization of the original metrics and leads to a faster learning process than the original framework. We also typically find that our ADDR term is most efficient in the original SCAN model than VSRN and BFAN. This may be due to VSRN and BFAN being developed to capture dense correspondence between visual region features, and their capability to adapt the features from different domains is originally stronger.

Figure \ref{examples} shows some examples of MSCOCO. In the sentence retrieval examples, it shows some discriminative textual concepts between the positive and negative sentences where we find they are highly correlated to the input images. Similarly, in the image retrieval results, we also find the difference between positive and negative images closely related to the phrase we highlight. These results confirm the capability of our ADDR models to compare and select the discriminative information when training with constructive samples.

Table \ref{mscoco} shows the experimental results on MS-COCO datasets. It compares the ensemble results of 2 of our single models on both 1k with 5-folder cross-validation and 5k test sets, by calculating the sum of their similarity scores. Comparing these sums shows that all of our ADDR models improve the initial baselines reported in the respective papers. On the 1k test set, our ADDR-VSRN ensemble outperforms the more recent published works by approximately $2\%$. More impressively, our ADDR-SCAN achieves $13.6\%$ better result on R@1 score and $5\%$ improvement on overall performance on the 5k test set. It is the best result among the existing works.

We present in Table \ref{flickr30k} the retrieval results on Flickr30K. As in the MS-COCO study, we also compare the retrieval performance of ensemble models on the test set with the same architecture settings. We also observe that our methods significantly outperform the other techniques. This indicates that our ADDR terms are also effective on the Flickr30K dataset. Comparing the columns in both Table \ref{mscoco} and Table \ref{flickr30k}, we also observe that our regularization terms help to improve more on R@1 than on other scores (e.g., our best result on 5k test set outperforms the best-reported results by about $7\%$ on R@1 score while only approximately $2\%$ to $4\%$ on R@5 and R@10, respectively). This may be due to the risk of R@1 most likely caused by the hard negative samples, while the similarity metric has already handled well on extended space of retrieval. Note that we learn to align the discriminate feature distributions in which the hard negative pairs are exclusively learned to be distinguished. We believe this is an interesting assumption for further exploration.

\vspace{-5pt}
\section{Conclusion}
\label{conclusion}
In this paper, we investigate the cross-modality problem by aligning the data distributions according to their semantic meanings. We propose an adversarial learning framework that externally regularizes the feature distribution within and between the sample pairs beyond the original metric learning. Our experiments confirm that our framework effectively aligns the data distributions to better support metric learning. It delivers impressive results showing competitive performance in the cross-modal retrieval task on the popular MS-COCO and Flickr30K datasets. We believe our adversarial regularization term can be easily extended to other similarity metrics and cross-modal representation learning of problems such as general information retrieval or image-based localization. We also intend to leverage the discriminative domain learning to improve the visual caption and scene generation systems in our future work.

\vspace{-5pt}
\bibliographystyle{IEEEtranS.bst}
\bibliography{submission-bib}

\end{document}